\documentclass{svproc}
\usepackage{url}
\usepackage{graphicx}
 \usepackage[breaklinks]{hyperref}
\usepackage{hyperref}
\usepackage{amsmath}
\usepackage[flushleft]{threeparttable}
\usepackage{fixfoot,xspace}
\usepackage[table,dvipsnames]{xcolor}
\usepackage[export]{adjustbox}
\usepackage{float}

\DeclareFixedFootnote*{\foo}{The experiment has just one run}
\begin{document}
\mainmatter             
\title{Global memory transformer for processing long documents}
\titlerunning{Global memory transformer for processing long documents}  

\author{Arij Al Adel\inst{1} 
\authorrunning{Arij Aladel et al.} 
%
\institute{Moscow Institute of Physics and Technology, Dolgoprudny, Russia\\
\email{arij.aladel@gmail.com}
}
}
\maketitle              

\begin{abstract}
Transformer variants dominate the state of the art in different natural language processing tasks such as translation, reading comprehension and summarization. Our paper is more directed to use general memory slots added to the inputs and studying the results of adding these slots.
This paper is a go on study of general memory slots rule that were added to the input of the proposed model in previous work \cite{AlAdel2021MemoryTW}. We have two main tasks;1) pretraining task using masked language modeling and b) fine tuning task using HotpotQA . This study aims to  verify the ability of the proposed model to handle chunks as if they were one chunk comparing with the base model. 
As baseline we used T5 transformer. We studied the rule of memory slots augmented to each input chunk and studied the model performance without selector. We found that adding memory to input chunks helped the proposed model to overcome the baseline on Masked language modeling task with specific training parameters. Ablation study reveals the ability of using the compressed input chunks with a degradation in performance.
\keywords{memory, transformer, attention, long documents}
\end{abstract}
\section{Introduction}
The power of processing long inputs appears clearly in NLP tasks like multi-hop and open-domain question answering, document classification, summarization, scientific paper summarization, document-level or multi-document relationship extraction , long documents translation, and coreference resolution tasks.
The quadratic complexity of the attention in transformer has always been the major transformers deadlock of scaling long inputs.
There were a series of works that used transformers with different kind of sparse attentions with additional global memory \cite{Ainslie2020ETCEL,DBLP:journals/corr/abs-2111-12763,DBLP:journals/corr/abs-2005-00979,DBLP:journals/corr/abs-1906-07651,Gupta2020GMATGM}.

Although all previous works have suggested really effective attention mechanisms still processing long documents is a challenging endeavor for many reasons. Concentrating on sequence to sequence models \cite{Beltagy2020LongformerTL,Guo2021LongT5ET,Gupta2020GMATGM} these challenges are not related solely to the encoder part but also to the decoder part. If the model were able to encode long input effectively there is still an obstacle to generate long responses in case of long dialogues, multi-document summarization, and story generation \cite{Guo2021LongT5ET}. This is because the same long input representation is used after encoding and passed to the decoder which makes the cross attention computationally expensive. Also most of tasks are still confined to the scope of classification long documents or generating short responses such as yes no answers.

This paper introduces intermediate results of on going work \cite{AlAdel2021MemoryTW} ; transformer variant with augmented memory slot prefixed into input chunks to tackle the chunked inputs processing as a step forward to process long documents. Last part introduces an ablation study of the model. We proposed using the memory slot of each chunk as a compressed vector presentation for each chunk into the encoder decoder attention of this model. Results reveals the ability to use the memory as compressed representation of the original chunks but with a degradation in performance. This is promising and can be one of future work directions.

In this model We abandon the usage of encoder output replacing it with selector output\cite{AlAdel2021MemoryTW}. This brings two advantages. First, more focus and attention on both source and target tokens during generation. Second, depending on the most relevant part from the input to generate the token paying attention to the overall input during generation. The main difference between our model and models\cite{Guo2021LongT5ET,Beltagy2020LongformerTL} that used the sparse attention is that chunking the input is happening inside the attention layer whilst input chunking in our model is already done in encoder before using the attention layer, which makes it easier to implement. We have two main tasks; we use MLM pretraining task with the same objective used in \cite{raffel2019exploring} and use question answering finetuning task on HotpotQA.

This study does not aim to benchmark on any of the mentioned data sets in this paper, but to verify the ability of the proposed model to handle chunks as
if they were one chunk comparing with the base model as a step forward to increase the input length.  Our work is following the line of works that added general memory to the transformer input \cite{Guo2021LongT5ET,AlAdel2021MemoryTW,Gupta2020GMATGM} and uses T5 as baseline to compare results and as a base structure to implement all the modifications.
\section{Global slot memory augmented Transformer with hierarchical attention }
\begin{figure}[H]
    \centering
        \centering
        \includegraphics[width=0.9\textwidth]{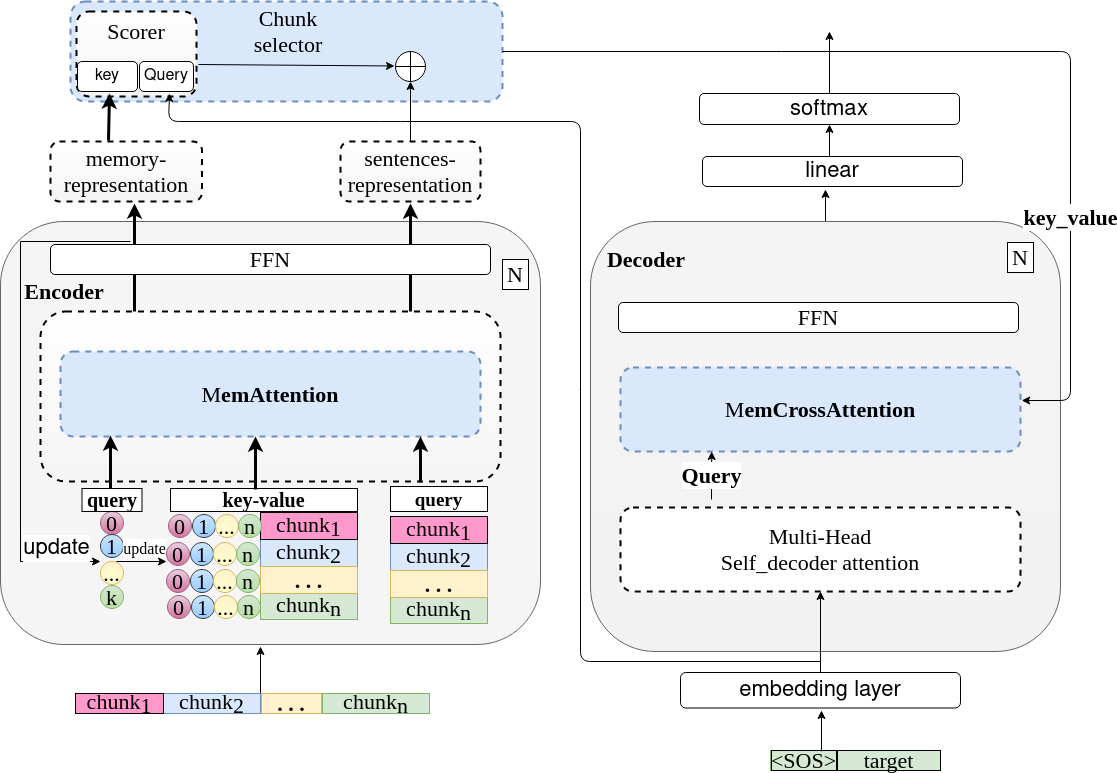} 
        \caption{Model overview with augmented memory slots in encoder part and chunk selector between encoder and decoder.}
        \label{fig:model}
\end{figure}

\begin{figure}[H]
    \centering
    \begin{minipage}{0.9\textwidth}
        \centering
        \includegraphics[width=0.9\textwidth]{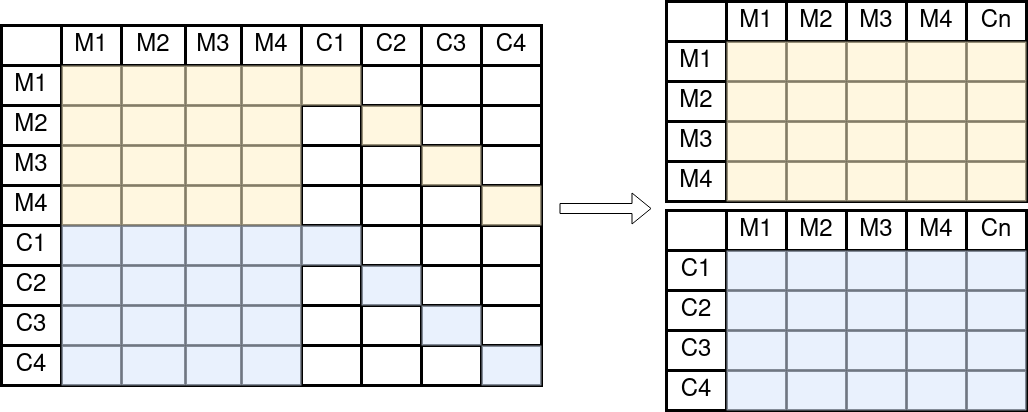} 
        \caption{Example attention pattern for handling input chunks in which dense attention is used for chunk.}
        \label{fig:attention}
    \end{minipage}
\end{figure}

    In light of our previous work and following the design steps of \cite{AlAdel2021MemoryTW} and depending only on just the forth design variant we experiment in this work.
    Fig \ref{fig:model} presents the overall model structure and fig.\ref{fig:attention} demonstrates how the implemented attention complexity and  the input chunks number are inversely proportional variables and the complexity becomes O(M · (L/n + M)) + O(L/n · (L/n+M)); L input length, M memory slot length where M$<<$L. Our choice of model design is related to the training time of the model. The model uses separate $W^{Q}$ projection matrix for memory slots and related chunks and shared $W^{K}, W^{V}$ in MemAttention.

\section{Masked language modeling}
For this task a PyTorch pipeline for masking tokens and building T5 objective style following \cite{raffel2019exploring} was implemented. The used data set is wikitext-103-raw-v1 \footnote{dataset can be found  \url{https://huggingface.co/datasets/wikitext/viewer/wikitext-103-raw-v1/train}}.

\subsection{Experiments setup} \label{MLM-setup}
We used trained tokenizer for all experiments. Tokenizer was trained following the steps of huggingface examples \footnote{steps of training the tokenizer can be found here \url{https://github.com/huggingface/transformers/tree/main/examples/flax/language-modeling\#train-tokenizer-2}}on training data files for both data sets of pretraining and finetuning tasks.
Since experimentally I found that using this trained tokenizer gives better results than using pretrained t5 tokenizer. This tokenizer was used for all experiments in this paper.
Vocabulary size is 32000 tokens for all experiments. We use batch size of 160 for all experiments during training and evaluating. Ada-factor optimizer was used and learning rate warmup over the first 2000 steps, linear decay of the learning rate with maximum linear learning rate \(lr = 0.005\). \(d_{model} = 512,d_k = 512/8 = 64, h = 8, num\_layers =2, d_{ff} = 2048\). 
For inference greedy search was used for all experiments. Training was done for 100 epochs. Drop out rate 0.1 during training and validation, seed= 42. Number of memory tokens for each memory slot is 2 tokens else mentioned otherwise. All experiments were averaged for two runs else noted otherwise under the table.

\subsection{Experimental MLM task results} 
We compared the results of the T5Mem model with T5(baseline) with two main input length a) input length 128 table \ref{tab:tab1} model uses just one chunk and b) input length 512 table \ref{tab:tab2} either as one chunk or divided into four chunks each chunk of 128 length. As can be seen in table \ref{tab:tab1} T5Mem model surpasses the base line. Table \ref{tab:tab2} T5Mem model surpasses the base line for both two variants using just one chunk or using four chunks. Anyway still results using just one chunk is better than using four chunks. Increasing input length while pretraining did not bring any good, and lower number of chunks with shorter length were more efficient than longer ones. In result, The proposed model using four chunks or one chunk outperformed the base line for both 128 input length as one chunk, 512 input length as one chunk, or 512 input length divided into four chunks each chunk of 128 length as seen in tables \ref{tab:tab1} and \ref{tab:tab2}.

    \begin{table}[H]
        \centering
        \caption{Experimental results on train and dev set of wikitext-103-raw-v1 dataset for Masked language modeling task using just one chunk as input length 128. Best results for the proposed model are highlighted.}\label{tab:tab1}

        \begin{threeparttable}
        \begin{tabular}{lcccc}
            \hline
            \textbf{Expriment name} & \textbf{Train loss}  & \textbf{Valid loss} & \multicolumn{1}{l}\textbf{Perplexity}\\
            \hline
            \textbf{T5\_basline\_128} &3.709 &3.455&32.265 \\
            \textbf{T5Mem\_MLM\_1\_chunk\_128} & \textbf{3.122} & \textbf{3.417} & \textbf{22.78} \\
            \hline
        \end{tabular}
        \end{threeparttable}
        
    \end{table}

    \begin{table}[H]
        \centering
        \caption{Experimental results on train and dev set of wikitext-103-raw-v1 dataset for Masked language modeling task using input length 512. Input length for all models is 512 length as one chunk or divided into four chunks of 128 chunk length. 128\_to\_512 means continue pre-training the model using 512 input length after training it for 100 epoches using input length 128 since the model is resilient to the input length.}\label{tab:tab2}
        \begin{threeparttable}
        \begin{tabular}{lcccc}
                \hline
                \textbf{Experiment name} & \textbf{Train loss}  & \textbf{Valid loss} & \multicolumn{1}{l}\textbf{Perplexity}\\
                \hline
                \textbf{T5\_basline\_512} & 4.327 & 4.204 & 66.94 \\
                \textbf{T5\_128\_to\_512} & 4.772 & 4.589 & 98.46 \\
                \textbf{T5Mem\_MLM\_1\_chunk\_512} & \textbf{4.101} & \textbf{4.186} & \textbf{60.422} \\
                \textbf{T5Mem\_MLM\_1\_to\_4\_chunks\tnote{a}  } & 4.735 & 4.5147 & 91.35 \\
                \textbf{T5Mem\_MLM\_4\_chunks\tnote{a}  } & \textbf{4.281} & \textbf{4.184} & \textbf{65.66} \\
                \hline
        \end{tabular}
        \begin{tablenotes}
            \item[a] Model has been trained once.
        \end{tablenotes}
        \end{threeparttable}
    \end{table}
    
\section{Fine tuning - question answering task}
For this task we have used HotpotQA dataset. HotpotQA \footnote{Read for more details about the dataset: \url{https://arxiv.org/pdf/1809.09600.pdf}}  is challenging data  set. It was introduced to encourage systems to learn more complex reasoning where the pieces of evidence to answer a question are scattered among different documents.
Gold paragraphs, i.e. the paragraphs containing the support,  facts  are included in the context paragraphs of distractor validation data set, whilst these gold paragraphs are not included in the context paragraphs in full wiki validation data set \cite{Yang2018HotpotQAAD}; for this reason the full wiki is more challenging data set.
HotpotQA  contains the  features: ['id', 'question', 'answer', 'type', 'level', 'supporting\_facts', 'context'] and has 90447 training record and 7405 validation record. Each context consists of a list of ten paragraphs. Each paragraph is a list of sentences. For each paragraph all sentences were merged in one text, then for each question all paragraphs were merged to form one context.
Supporting facts were not used. We have used two setups for input length. First one using source length of 512 tokens, target of 40 tokens length and one chunk. Second one using source length of 512 tokens divided into four chunks, target of 40 tokens length to verify the ability of the proposed model of processing chunked input as one input. The total input length is always equal to num\_chunks multiplied by source\_length. 
I have used distractor data set. 
We have used both baseline and the proposed model as one stage sequence-to-sequence models to get the answer of the input that has the following structure: {question\_ids $<$/s$>$ context\_ids $<$/s$>$}. target as {target\_ids$</s>$}. For the other experiments setups stay the same as for MLM experiments setup in the section \ref{MLM-setup}. Results of finetuning showed that both the baseline T5 and the proposed models did not manage to tackle the task(all metrics were zeros). That is why we tried simply to use fixed learning rate; this worked and model started to tackle the task. Table \ref{tab: tab5} shows the results and base line overcomes the proposed model. 
\section{Ablation study}
\begin{figure}[H]
            \includegraphics[width=0.9 \textwidth]{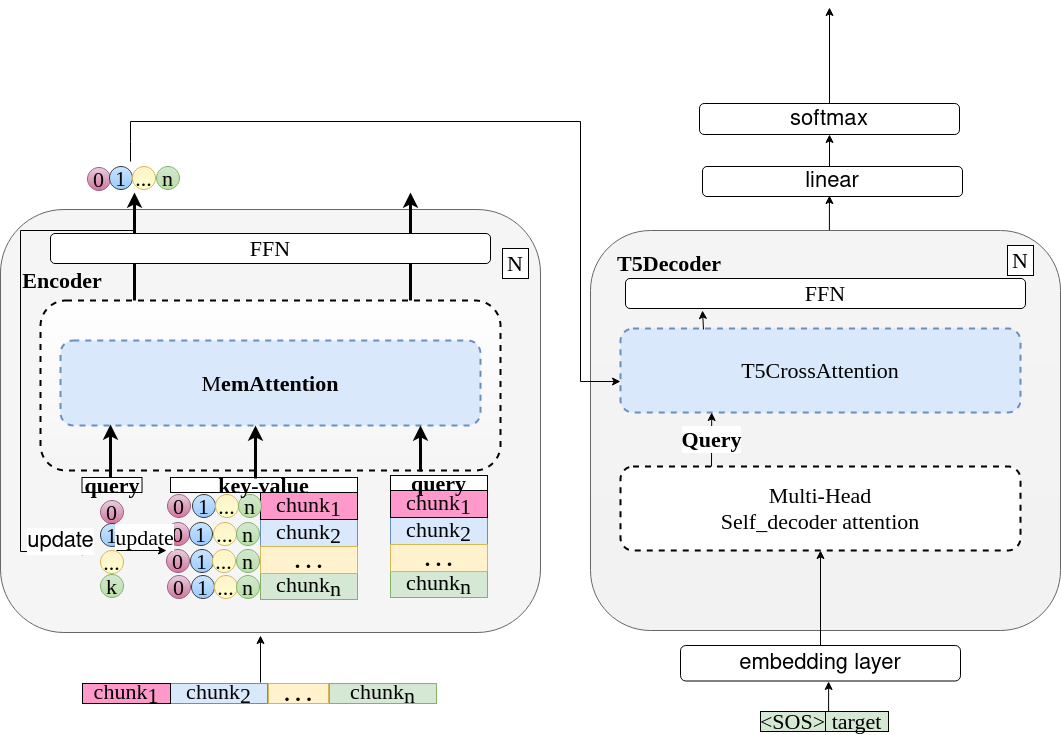}
            \caption{T5MemWS the modified proposed model with dropped selector(WS=without selector)}
            \label{fig:sentence_transformer_ws}
        \end{figure}

\begin{figure}[H]
            \includegraphics[width=0.9 \textwidth]{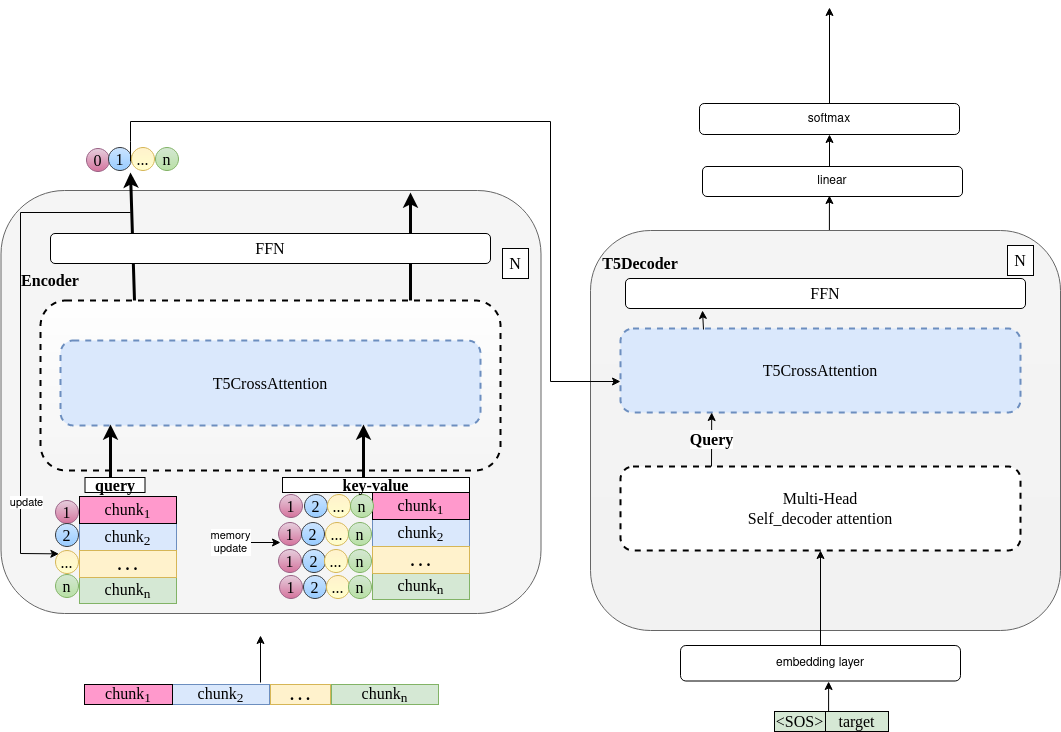}
            \caption{T5MemWs\_WMA modifying proposed model without selector and T5cross attention in encoder}
            \label{fig:sentence_transformer_ws_cr}
        \end{figure}

To verify to which degree the model benefits from the selector we drop the selector. We have used the memory slots directly after encoder with two additional modifications: a) simply drop out the selector and use the concatenation of memory slots for all input chunks directly to decoder cross attention fig.\ref{fig:sentence_transformer_ws} , b) drop out the selector and use the concatenation of memory slots for all input chunks directly to decoder cross attention, but in encoder we use T5cross attention between memory and chunk representations fig.\ref{fig:sentence_transformer_ws_cr} . 
This model fig.\ref{fig:sentence_transformer_ws} comes for two reasons; a) reduce the computational cost of the encoder decoder attention in our proposed model b) check using the memory slots as valid chunk representation (compressed representation). There were not any kind of hierarchical attention over the memory slots here since each memory token in each slot has access to all other memory slots tokens in encoder attention. In this design variant  of the model the mathematical representation of the model still the same as mentioned in \cite{AlAdel2021MemoryTW}. Equations of selector are dropped. Equations of encoder and decoder stay the same for causal attention except the equations for decoder cross attention become:
\begin{gather}
    \scalebox{0.9}{$att = LN(X^{Cusal\_output}) + T5Attention(X^{Cusal\_output}, X^{mem}_{all\_chunks}, X^{mem}_{all\_chunks})$}
    \label{eq:memcross}
\end{gather}
 T5 positional encoding style was preserved in all models. Experiments setup stay the same as\ref{MLM-setup} except we experiment with additional optimizer Adam (in tables Adm) and use also constant learning rate $lr=5\times 10^{-5}$ results in tables \ref{tab:tab3}, \ref{tab:tab4} and table \ref{tab: tab5}. In the case of linear optimizer we use learning rate $lr=3\times10^{-3}$ warm up setps for all experiments were 2000 steps. To read the name of experiments right in the tables \ref{tab:tab3} and \ref{tab:tab4}, we put the name of used model (T5, T5Mem, T5MemWs\_2mem, T5MemWsWMA\_2mem, or T5MemWsWMA\_1mem);
 T5Mem the main name of the proposed model, Ws means without selector, WMA means without memory attention so the structure of name is:\\
 {Model}\_{optimizer}\_{scheduler}
\subsection{Ablation study results}
Results of experiments on these two new variants compared with the previous results presented in the tables \ref{tab:tab3} and \ref{tab:tab4}. As we notice in table \ref{tab:tab4} using constant scheduler during pretraining has strong effect on the pretraining process. Baseline using constant scheduler was better than all proposed model versions. Using MemAttention is better than using T5Cross attention regarding all results, which emphasises superiority of MemAttenion on T5Attention. Using separate parameters for memory slot and its related chunk helped to get this improvement.  We have tried to use just one memory token for each chunk in T5MemWS\_WMA\_1mem model we noticed that the quality of results were decreased using Adam optimizer. Which emphasizes that increasing memory slot capacity can improve the results. Base line using Adam optimizer and constant learning rate gives better accuracy but not better perplexity comparing with itself. While for our model T5Mem and all its modifications using Adam optimizer with constant learning rate led to worse results regarding both accuracy and perplexity. 
Table \ref{tab: tab5} displays the results of finetuning the pretrained models from table \ref{tab:tab3} and \ref{tab:tab4} on question answering task using hotpotQA dataset. The same tendency of using constant learning rate to improve the results was kept during finetuning. Still baseline is better than the proposed model but both the baseline and the proposed model in all design variants start tackling the task. What is interesting in fientuning is that results with dropped selector and MemAttention used were the best comparing with all other model variants except for (T5Mem. AF,const) which was the best between all model variant results. Using constant learning rate no optimizer gives better results on all experiments. Results of finetuning again display preference MemAttention on T5cross attention and the role of compressed representation in auto regressive response generation. The length used in these experiments is not long enough to see if for longer inputs the model will do better, this is possible in the future work. In the result we understand that the selector does the mission using linear learning rate and surpass the baseline but the performance is on par during finetuning with the baseline. While using constant learning rate enables the models to handle the finetuning task but the baseline is better for all the experiments that used constant learning rate. Using compressed representation for extractive responses is possible and promising. This can mitigate the expensive computational cost of the cross attention in the decoder while processing long documents. 

\begin{table}[H]
    \centering
    \caption{Experimental MLM task results using just 2 layers - training for 100 epochs input length is 512 for T5. For T5Mem model and its variants the input length 512 is divided into 4 chunks. optimizers(Adafactor ``AF`` and AdamW ``Adm``), the used scheduler is \textbf{linear}. All models use two memory tokens for each chunk as a slot, except for T5MemWsWMA\_1mem uses 1 memory token. Lines in grey use Adam optimizer. Best results for Adafactor optimizer are underlined. Best results for Adam optimizer are bold.}

    \label{tab:tab3}
    \fontsize{9}{11}
    \setlength\tabcolsep{0pt}
    \begin{threeparttable}
    \begin{adjustbox}{max width=\textwidth}
    \begin{tabular}{lccccccc}
        \hline
        Model                           & Train loss. & Valid loss. & val\_acc. & {val\_PPL} & Test loss. & test\_acc. & \multicolumn{1}{l}{test\_PPL} \\ \hline
        T5. AF,linear\tnote{1}          & 4.327  & 4.196   &  30.060   & 66.394  & 4.163  &  30.218  & 64.289 \\
        \rowcolor{black!10}
       \rowcolor{black!10}
        T5. Adm,linear                  & 4.368  & 4.258   & 27.605   & 70.648  & 4.234 & 27.675 & 68.991 \\ \hline
        T5Mem. AF, linear\tnote{2,3}    & 4.286  & 4.171   &  29.775  & 64.769  & 4.140  &  29.808  & 62.816 \\
        \rowcolor{black!10}
        \rowcolor{black!10}
        T5Mem Adm,linear\tnote{2}       & 4.422  & 4.301   & 27.620    & 73.754  & 4.280   & 27.919   & 72.258 \\\hline
        T5MemWs\_2mem. AF,linear         & 4.193 & 4.093 & 30.090 & 59.979 & 4.091 & 30.891 & 59.779 \\

        \rowcolor{black!10}
        \rowcolor{black!10}  
         T5MemWs\_2mem. Adm,linear      & 4.247 & 4.143 & \textbf{30.210} & \textbf{62.998} & 4.113 & \textbf{30.397} & \textbf{61.158}\\\hline
        
        T5MemWsWMA\_1mem. AF,linear     &  4.149 &  4.050  &   \underline{31.145}  &  \underline{57.417}& 3.999   & \underline{31.355}   & \underline{54.589} \\

        \rowcolor{black!10}
        T5MemWsWMA\_1mem. Adm,linear    &  4.307  &  4.202   &  28.316 &  66.873   &  4.176  & 28.316  &  65.117  \\ 

        T5MemWsWMA\_2mem. AF,linear     & 4.196  &  4.069  &  31.010   &  58.497 & 4.0217  &  31.262  & 55.813 \\

        \rowcolor{black!10}
        T5MemWsWMA\_2mem. Adm,linear    & 4.266  &  4.161  & 29.985   &  64.111 &  4.136   &  30.103 & 62.548 \\\hline

    \end{tabular}
    \end{adjustbox}
    \begin{tablenotes}
        \tiny
        \item[1] It is the model T5\_baseline\_512 the name in table just to make it easy to compare in this table.
        \item[2] Model has been trained once.
        \item[3] It is the model T5Mem\_MLM\_4\_chunks name was changed in the table to make it easy to compare
        
    \end{tablenotes}
            
    \end{threeparttable}
\end{table}
\begin{table}[H]
    \caption{Experimental MLM task results using just 2 layers - training for 100 epochs input length is 512 for T5. For T5Mem model and its variants the input length 512 is divided into 4 chunks. optimizers(Adafactor``AF`` and AdamW ``Ad``), the used scheduler is constant. All models use two memory tokens for each chunk as a slot, except for T5MemWsWMA\_1mem uses 1 memory token. Lines in grey use Adam optimizer. Best results for Adafactor optimizer are underlined. Best results for Adam optimizer are bold. For all best results baseline is superior.}
    \label{tab:tab4}
    \fontsize{9}{11}
    \begin{threeparttable}
    \centering
    \resizebox{\textwidth}{!}{
    \begin{tabular}{lccccccc}
        \hline
        Model                           & Train loss. & Valid loss.   & val\_acc. & {val\_PPL} & Test loss. & test\_acc. & \multicolumn{1}{l}{test\_PPL} \\ \hline
        T5. AF,const                    & 2.486  & 2.308  &  59.225  & 10.0595   & 2.265  &  60.150           & 9.634 \\
        \rowcolor{black!10}
        T5. Adm,const                   & 2.495  & 2.309   &  59.59   & 10.060    & 2.269  &  60.523           & 9.673 \\ 
        T5Mem. AF,const\tnote{a}        &  3.094 & 2.892   &  \underline{52.47}   & 18.020    & 2.864  & \underline{52.742}  & \underline{17.533} \\
        \rowcolor{black!10}
        T5Mem. Adm,const\tnote{a}       & 3.304  & 3.078   &  47.77   & 21.720    & 3.757   & 33.659           & 42.822 \\
        T5MemWs\_2mem. AF,const         & 3.126  &  2.889  & 50.87    & \underline{17.990}    & 2.865   & 51.175           & 17.555 \\

        \rowcolor{black!10}
        T5MemWs\_2mem. Adm,const        & 3.117  & 2.886   & \textbf{51.005}   & \textbf{17.916}    & 2.864   & \textbf{51.213}           & \textbf{17.531} \\
        T5MemWsWMA\_1mem. AF,const      & 3.369  &  3.157 &  44.805  & 23.495    & 3.126   &45.216           & 22.789 \\

        \rowcolor{black!10}
        T5MemWsWMA\_1mem. Adm,const     & 3.328 & 3.104  & 47.275   & 22.290     & 3.076  &  47.656          & 21.677 \\ 

        T5MemWsWMA\_2mem. AF,const      & 3.280 &  3.086  &  47.585  & 21.541     & 3.055   &  47.961          & 21.225 \\

        \rowcolor{black!10}
        T5MemWsWMA\_2mem. Adm,const     & 3.277 & 3.085   & 47.465   & 21.88      & 3.0535  & 47.806           & 21.195 \\

    \end{tabular}
    }
        \begin{tablenotes}
        \tiny
            \item[a] Model has been trained once.
        \end{tablenotes}
        
    \end{threeparttable}
\end{table}
\begin{table}[H]
    \caption{Experimental results of fine tuning all pre-trained models and modified models on HotpotQA dataset(distractor part) using input length as 512 even as one chunk(T5 take input as chunk of 512 length) or 512 input length is divide it to 4 chunks. Lines in grey use Adam optimizer. All experiments were pretrained and finetuned using constant learning rate}\label{tab: tab5}
   \begin{threeparttable}
        \centering
        \resizebox{\textwidth}{!}{
        \begin{tabular}{lcccccccc}
            \hline
            \textbf{Experiment name} & \textbf{Train loss}  & \textbf{Valid loss}& \textbf{Test loss}& \textbf{exact match}& \textbf{f1}& \textbf{recall}& \multicolumn{1}{l} \textbf{precission}\\
            \hline
            \textbf{T5\_hp\_AF\_const\_512} & 0.757 & 3.231 & 3.231 & 17.008 & 24.619 & 24.995 &  25.825 \\
            \rowcolor{black!10}
            \textbf{T5\_hp\_Adm\_const\_512} & 0.755 & 3.285 & 3.285 & 16.819 & 24.175 & 24.522 & 25.279 \\
            \textbf{T5Mem\_hp\_4\_chunks\_AF\_const} & 1.696 & 4.024 & 4.024 & 8.727 & 14.007 & 13.977  &  14.894\\
            \rowcolor{black!10}
            \textbf{T5Mem\_hp\_4\_chunks\_Adm\_const} & 2.416 & 4.150 & 4.150 & 4.090& 5.507 & 5.480  & 5.781 \\
            \hline
            \textbf{T5MemWS2mem\_hp\_4\_chunks\_AF\_const} & 1.227 & 4.090 & 4.090 & 7.930 & 14.227 & 14.190  & 15.190 \\
            \rowcolor{black!10}
            \textbf{T5MemWS2mem\_hp\_4\_chunks\_Adm\_const} & 1.236 & 4.133 & 4.133 & 8.140 & 14.797 & 14.369  & 15.372 \\
            \hline
            \textbf{T5MemWSWMA1mem\_hp\_4\_chunks\_AF\_const} & 1.296 & 4.052 & 4.052 & 7.876 & 13.905 & 13.911  & 14.904 \\
            \rowcolor{black!10}
            \textbf{T5MemWSWMA1mem\_hp\_4\_chunks\_Adm\_const} & 1.112 & 4.331 & 4.331 & 6.965 & 12.657 & 12.610  & 13.587 \\
            \textbf{T5MemWSWMA2mem\_hp\_4\_chunks\_AF\_const} & 1.233 & 4.217 & 4.218 & 7.843 & 13.964 & 13.966  & 14.87 \\
            \rowcolor{black!10}
            \textbf{T5MemWSWMA2mem\_hp\_4\_chunks\_Adm\_const} & 1.117 & 4.401 & 4.401 & 6.959 & 12.888 & 12.941  & 13.658 \\
            \hline
            
        \end{tabular}}
    \end{threeparttable}
\end{table}

\section{Conclusion and future work}
This paper presented a study of our proposed model on two new tasks; Masked language modeling as pretraining task and question answering task using HotpotQA as finetuning task. Experimental results showed good performance of the model on masked language modeling using linear learning rate. The prposed model with chunked input outperformed T5 as baseline. This approves previous results on translation task where the proposed model overcomes the base line. Results of the proposed model were on par with the baseline on fintuning task where both models do not manage to solve the finetuning task.There were not preferred optimizer results between two different optimizers adam and adafactor. Using constant learning rate the proposed model does not demonstrate good results for chunked input using the chunk selector for funetuning on hotpotQA .On the other side, using memory slots as compressed representation of chunks led to better results on funetuning task than the proposed model but did not beat the baseline T5 yet. This is interesting results that can be studied more in the future work in encoder decoder transformer as T5, since all previous sequence to sequence works still use the long input representation into the decoder part.
    

\bibliographystyle{styles/bibtex/splncs03}
\bibliography{references.bib}

\end{document}